\pdfoutput=1

\documentclass[11pt]{article}

\usepackage[preprint]{acl}

\usepackage{times}
\usepackage{latexsym}

\usepackage[T1]{fontenc}

\usepackage[utf8]{inputenc}

\usepackage{microtype}
\usepackage{inconsolata}

\usepackage {amsmath}
\usepackage {amssymb}
\usepackage {bm}
\usepackage {bibnames}
\usepackage {graphicx}
\usepackage {graphpap}
\usepackage {longtable}
\usepackage {multirow}
\usepackage {tikz}
\usepackage{booktabs}
\usepackage{xcolor}
\usepackage{xspace}
\usepackage{natbib}
\usepackage{cleveref}

\usepackage{makecell}
\usepackage{array}
\usepackage{amsthm}
\usepackage[normalem]{ulem}

\newcommand{\qwen}{Qwen-2.5\xspace}

\title{Test-Time Scaling with Repeated Sampling Improves\\ Multilingual Text Generation}

\author{
\begin{tabular}{c}
Ashim Gupta \quad\quad\quad Vivek Srikumar
\end{tabular}\\
Kahlert School of Computing \\
University of Utah \\
\texttt{ashim@cs.utah.edu}}

\begin{document}
\maketitle
\begin{abstract}
Inference-time scaling via repeated sampling has shown promise in reasoning tasks, but its effectiveness in multilingual generation remains underexplored. We evaluate this approach using perplexity- and reward-based verifiers on two multilingual benchmarks: the Aya Evaluation Suite and m-ArenaHard. Our results show consistent quality improvements, with gains exceeding 35\% in some cases.
While perplexity-based scoring is effective for open-ended prompts, only reward-based verifiers improve performance on tasks requiring reasoning (e.g., math, code). Our results demonstrate the broader utility of repeated sampling for multilingual text generation and underscore the importance of selecting right verifiers for the task. 
\end{abstract}

\section{Introduction}
%
%
%
%
Scaling the inference-time compute has significantly improved LLM performance on coding, mathematics, and other reasoning tasks ~\cite[e.g.,][]{madaan2023self,wu2024inference,snell2024scaling,brown2024large,bansal2024smaller,muennighoff2025s1}. There are two broad strategies for test-time scaling: (1) with lengthy chains of thought using an LLM trained to \textit{reason} longer~\citep{guo2025deepseek,muennighoff2025s1}, or (2) by repeated sampling, where a verifier or a scorer model selects a final answer from generated samples~\citep{brown2024large,snell2024scaling}.

\begin{figure}[t]
    \centering
    \includegraphics[width=0.99\columnwidth]{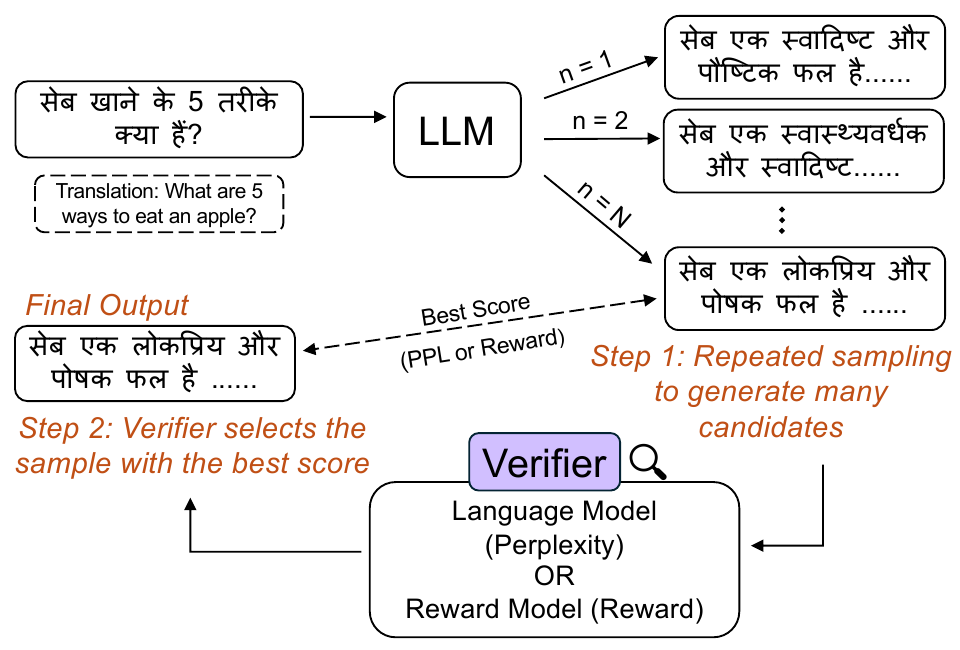}
    \caption{Repeated sampling procedure using a verifier to pick the final answer.}
    \label{fig:verifier_example}
\end{figure}

Despite the success of test-time scaling in reasoning-heavy tasks, its
effect on multilingual text generation
remains largely unexplored. In this work, we investigate whether the \textit{repeated sampling} strategy
can improve multilingual generation, even for benchmarks that do not explicitly focus on reasoning, code, or mathematics. We employ two approaches for selecting the final output: (a) perplexity-based scoring using an auto-regressive language model, and (b) reward scoring with pre-trained reward models. The perplexity-based method considers only the generated responses, selecting the output with the lowest perplexity score as a proxy for fluency. In contrast, the reward-based method evaluates both the prompt and the generated response together, scoring each prompt-response pair to determine the best output. 
\Cref{fig:verifier_example}  illustrates this procedure.


Repeated sampling may allow models to explore a wider range of potential outputs. Consequently, in the case of multilingual generation, especially for low-resource languages, the likelihood of generating text that is both high quality and contextually appropriate increases. 
When model confidence is limited, e.g., for underrepresented languages, initial outputs may not be the best. By selecting from multiple generated candidates using criteria based on fluency (via perplexity) or prompt-response alignment (via reward models), we aim to identify outputs that are more consistent with human judgments or benchmark standards.

We show empirically that repeated sampling with verification improves multilingual text generation. Using two perplexity- and reward-based verifiers each, we observe that output quality improves  with the number of samples. On the Aya Evaluation Suite~\citep{singh2024aya}, a multilingual benchmark for open-ended generation, both verifier types yield substantial gains; a 2B-parameter Gemma-based reward model achieves up to 35\% increase in win rates. However, for m-ArenaHard~\citep{dang2024aya}, a benchmark focused on  programming and mathematics, only reward-based verifiers lead to measurable improvements.

Our contributions are twofold. (1) Ours is the first work to demonstrate that multilingual generation benefits from test-time scaling, highlighting the broader applicability of repeated sampling beyond traditional English-based reasoning tasks. (2) We show that verifier choice is critical: both perplexity- and reward-based scoring helps general multilingual generation, but only reward-based verifiers are effective for tasks requiring reasoning.

\section{Experimental Setup}
This section outlines our experimental setup, including details on the datasets, languages, LLMs, verifiers, and the evaluation protocol used. 

\paragraph{Dataset and Languages.} We use two datasets for evaluation. First is the Aya Evaluation Suite~\cite{singh2024aya}, that contains open-ended conversation-style, culturally grounded prompts for evaluating multilingual LLM generation. We conduct evaluations on nine languages: 
Arabic, English, Hindi, Punjabi, Portuguese, Russian, Telugu, Turkish, and Chinese.

The second dataset is m-ArenaHard~\cite{dang2024aya}, the multilingual variant of ArenaHard-Auto~\citep{li2024crowdsourced}. This dataset contains the translated prompts from the ArenaHard-Auto benchmark. With this dataset, we perform evaluation in seven languages: 
Arabic, Czech, English, Hindi, Japanese, Portuguese, and Vietnamese. 

\paragraph{Verifiers.} 
We employ two types of verifiers to score and select generations from repeated sampling. For perplexity-based verification, we select sample that has the lowest perplexity and use the pre-trained variants of \texttt{LLaMA-3.1-8B}~\citep{grattafiori2024llama} and \texttt{Gemma-2B} models~\citep{gemma_2024}.
%
For reward-based scoring, we use two fine-tuned reward models that are based on the above pre-trained ones: \texttt{URM-LLaMA-3.1-8B}~\citep{lou2024uncertainty}, and \texttt{GRM-Gemma-2B-rewardmodel-ft}~\citep{yang2024regularizing}. 
Both reward models rank high on the RewardBench benchmark~\cite{lambert2024rewardbench} and are among the best models in their respective parameter ranges. 
Note that while the underlying pre-trained models are multilingual, the reward models have only been trained on the English preference data. 
\Cref{subsec:appendix:verifier} elaborates on our verifiers.
\paragraph{LLMs Evaluated.}
We evaluate commonly used  open-weight multilingual LLMs at both their large and small parameter scales. Specifically, we use Qwen-2.5~\cite{yang2024qwen2} at 14B, 72B parameter scales, Llama-3 at 8B and 70B scales~\cite{grattafiori2024llama}, and aya-expanse at 8B and 32B~\cite{dang2024aya}. 
\paragraph{Evaluation Protocol.}
We follow prior work and adopt the LLM-as-a-Judge framework for evaluating multilingual generation~\cite{dang2024aya}, based on MT-Bench~\cite{zheng2023judging}. The core evaluation metric is the win/loss rate\,---\,i.e., how often a model's response is judged better or worse than a baseline. We report the difference (\textit{delta}) between win and loss rates, averaged across all languages. All evaluations use \texttt{gemini-2.0-flash} as the judge, which we choose because of its cost efficiency and strong multilingual performance.

Our baseline consists of single-sample generations without repeated sampling, using temperature-based decoding. Due to its stochastic nature, the baseline output may vary across runs, introducing randomness into the delta scores. Comparing against multiple baseline samples is impractically expensive. In a pilot study (\cref{fig:stability}, appendix),  we found that the variance across runs is small ($<2$ percentage points), 
%
indicating that our findings are robust to baseline selection.

\begin{figure*}[ht]
    \centering
    \includegraphics[width=\textwidth]{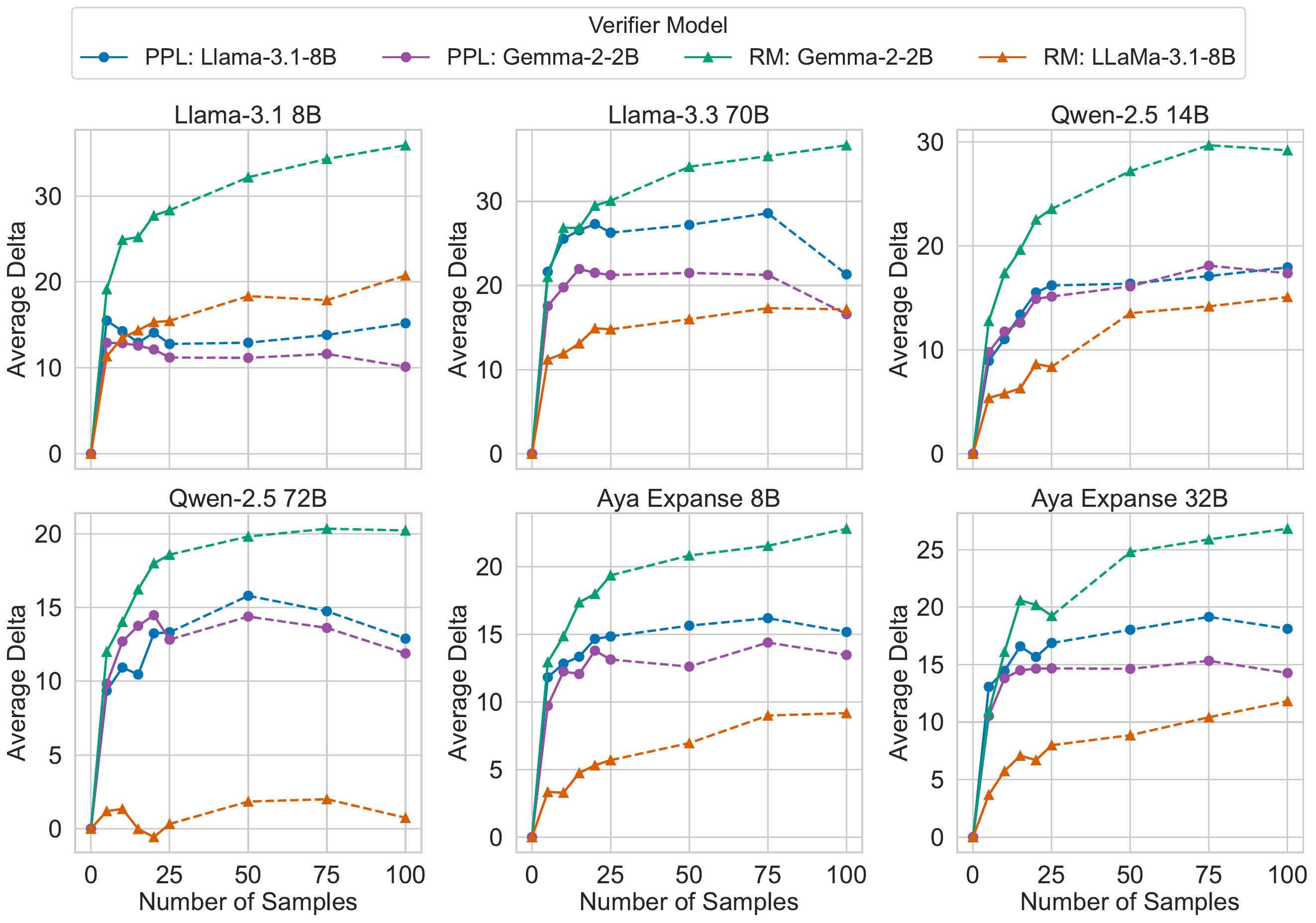}
    \caption{Test-time scaling with repeated sampling for Aya Evaluation Suite. The plots show the difference between win and loss rates (delta). We see that all verifiers\,---\,both perplexity-based (PPL) and reward-based (RM)\,---\,can improve generation quality.}
    \label{fig:scaling_main}
\end{figure*}
\section{Results and Analysis}
We plot the impact of number of samples ($n$) used for repeated sampling vs the average delta score achieved to show the scaling effect. We perform this evaluation for $n = 5, 10, 15, 20, 25$ and then scale it in steps of 25 after that (to $n = 50, 75, 100$). 

\begin{figure*}[ht]
    \centering
    \includegraphics[width=\textwidth, trim=2 2 0 0, clip]{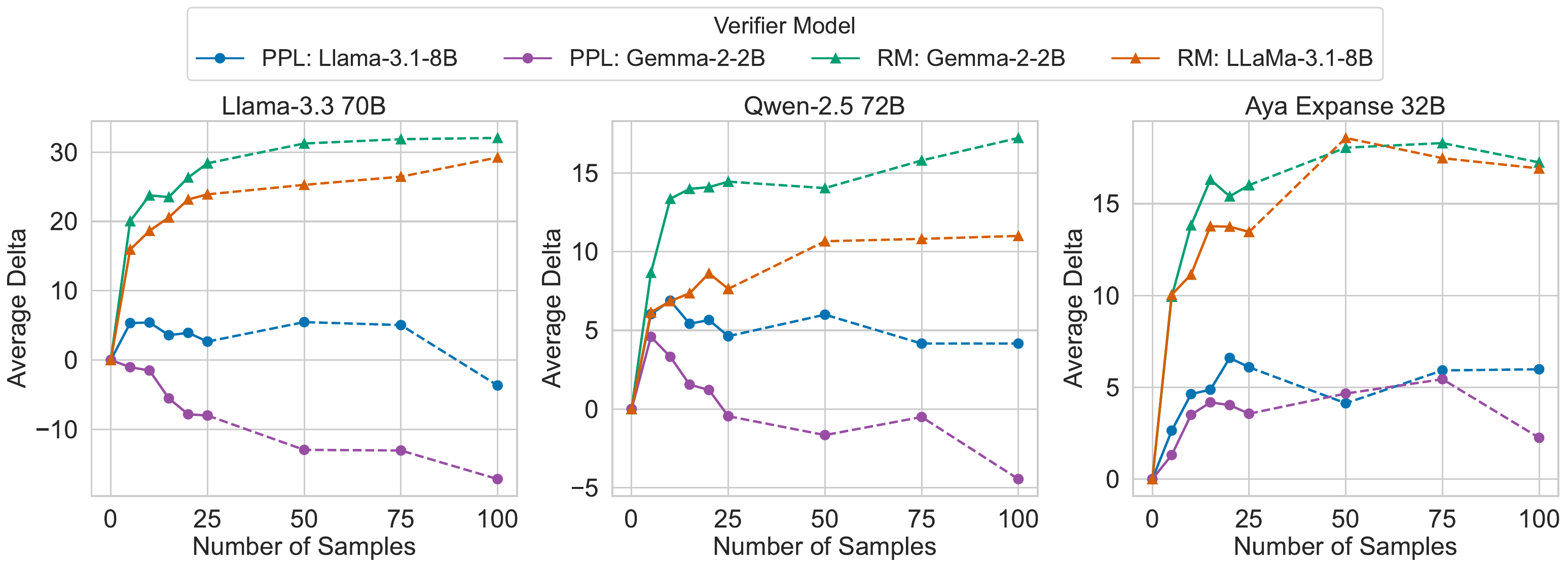}
    \caption{Test-time scaling with repeated sampling for m-ArenaHard. The plots show the difference between win and loss rates (delta). Only reward-model based verifiers improve generation quality.}
    \label{fig:scaling_main_arena_small}
\end{figure*}

\begin{figure}[t]
    \centering
    \includegraphics[width=0.92\columnwidth, trim=8 8 0 6, clip]{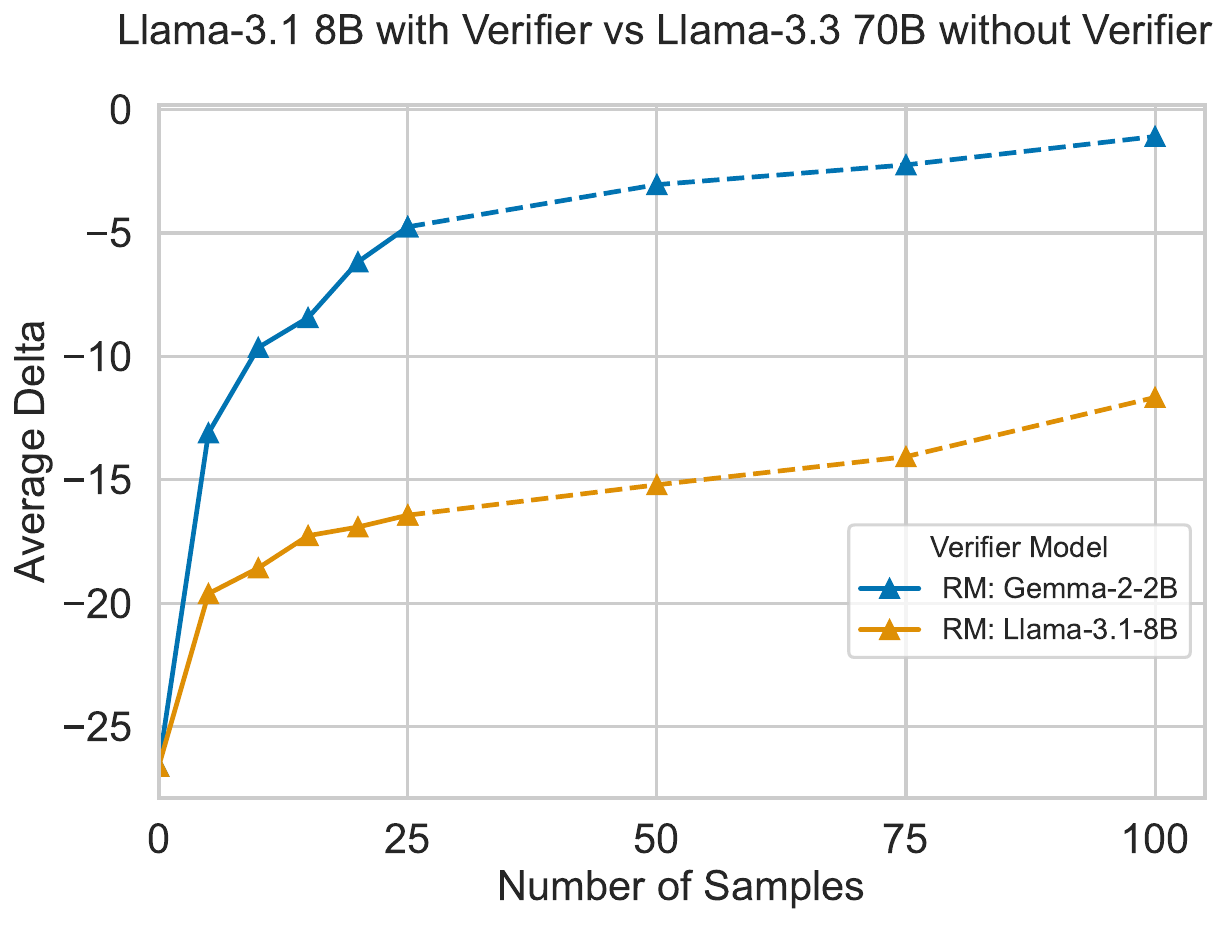}
    \caption{Training-time compute vs Test-time compute.}
    \label{fig:large_vs_small}
\end{figure}

\Cref{fig:scaling_main} shows the main scaling plots for  the Aya Evaluation Suite and \cref{fig:scaling_main_arena_small} for the m-ArenaHard.
For the latter, due to space constraints, we only show the plots for the larger models of each model family. Other plots are presented in appendix~\cref{fig:scaling_main_arena_all}. 

\paragraph{1. \uline{Repeated sampling with verifiers improves multilingual generation.}} For Aya Evaluation Suite, across all model architectures, verifier types, and scales, repeated sampling during inference  improves multilingual generation. This effect is evident even with a modest number of samples; we see significant performance gains as early as $n=5$. While improvements continue with more samples, the marginal benefit begins to plateau around $n=25$ for most models. An exception is the pairing of the \texttt{URM-LLaMA-3.1-8B} reward model with the~\qwen 72B model, where gains are minimal. Yet, the consistent upward trend supports the value of repeated sampling for improving generation quality in multilingual settings.

\noindent\textbf{Aya Evaluation Suite vs m-ArenaHard.} 
For m-ArenaHard, only reward-based verifiers consistently improve performance. In fact, perplexity-based verification can even hurt generation quality with more samples (e.g., 70B-parameter Llama-3 using Gemma verifier). This observation aligns with the nature of the tasks: open-ended textual prompts predominate the Aya Evaluation Suite, where fluency correlates strongly with quality. In contrast, m-ArenaHard prompts demand domain-specific reasoning in mathematics or programming, where surface-level fluency is insufficient. Reward models, trained extensively on reasoning-intensive math/coding datasets, are better at distinguishing high-quality responses in these settings.

\paragraph{2. \uline{Llama-based verifiers improve Llama-based models.}}
Interestingly, Llama-based verifiers can help select higher quality texts from their post-trained counterparts. 
This is  unexpected, as post-trained models are typically optimized to align with either their own reward signals or their internal fluency preferences, suggesting they might better self-evaluate. Nonetheless, Llama verifiers appear well-suited to identifying strong outputs from related model families.

\paragraph{3. \uline{Gemma-based reward model delivers the highest gains.}}
The reward model based on Gemma-2  delivers the highest gains across both benchmarks. At $n = 100$, all generation models improve by over 20\%, with Llama-8B and Llama-70B achieving increases exceeding 35\%. This is particularly notable given that the Gemma verifier has only 2B parameters. These results indicate that, under repeated sampling, even a relatively small reward model can substantially enhance the output quality of much larger language models. Importantly, the verifier was trained exclusively on English prompt-response data;  further improvements may be possible with multilingual training.

\paragraph{4. \uline{Verifier choice is critical.}} As observed on the m-ArenaHard benchmark, reward-based verifiers substantially enhance generation quality, whereas perplexity-based methods often degrade it. Thi highlights the importance of choosing the right verifier for the right task.  Moreover, verifier performance can vary across languages, suggesting that language-specific factors influence effectiveness. See for instance the heatmaps provided in appendix~\cref{fig:heatmap_aya} and~\cref{fig:heatmap_arena}. 

For example, while the Gemma-based perplexity verifier appears more effective for Qwen models on Hindi, a Llama-based verifier yields stronger results on Arabic. These variations suggest that no single verifier might be universally optimal, and future work could explore strategies for selecting verifiers based on language, model architecture, or task type, and training language-agnostic verifiers.
\paragraph{Train-time Scaling vs Test-time Scaling.} 
One question we ask is whether test-time scaling can compensate for the limitations of smaller models, potentially narrowing the gap with larger models trained using significantly more compute. To explore this, we compare the outputs of Llama-3 8B with repeated sampling (test-time scaling) against Llama-3 70B without sampling (train-time scaling only). As shown by the negative delta values in~\cref{fig:large_vs_small}, while repeated sampling does improve the smaller model's outputs, it does not match the quality achieved by the larger model, indicating that test-time scaling alone cannot fully substitute for increased training-time capacity. This suggests that while test-time scaling can enhance performance, especially for smaller models, it might not be a replacement for the representational capacity and capabilities gained through large-scale training.

\section{Conclusion}
This work establishes inference-time scaling via repeated sampling as a practical and effective strategy for improving multilingual text generation. Our analysis reveals that performance gains are robust across models, verifier types, and languages, with notable improvements achievable even for small-scale verifiers. Crucially, we find that verifier effectiveness can be task and language-dependent: perplexity-based methods work well for open-ended prompts but underperform on domains requiring structured reasoning. These findings highlight the importance of careful verifier selection and open up promising directions for future work, including adaptive strategies that tailor verifier choice to the characteristics of the input and the model.

\section*{Limitations}
We identify two primary limitations in this work. First, the reward models employed are trained solely on English data. This choice reflects the limited availability of compact, high-performing multilingual reward models. Existing open-weight multilingual reward models, such as those in m-RewardBench, tend to be prohibitively large (e.g., Gemma-2 at 27B), making them impractical as lightweight verifiers in repeated sampling setups. Future work could address this by developing smaller, more efficient multilingual reward models.

Second, due to budget constraints, our evaluation uses a single baseline sample when computing win-rates, which may introduce variance in the results. Although our analysis shows that this variance is relatively small, its impact across different models and languages remains uncertain. Moreover, our use of LLM-as-a-Judge introduces an additional layer of potential bias, as the judgments reflect the preferences and limitations of the underlying judge model. While this approach is practical and widely adopted, it may not fully align with human evaluations, particularly in multilingual or culturally nuanced contexts.


\section*{Acknowledgements}
We thank the members of the UtahNLP group for their constructive feedback. Ashim Gupta is supported by the Bloomberg Data Science Ph.D. Fellowship. This material is based in part upon work supported by the National Science Foundation under Grant \#2217154. The support and resources from the Center for High Performance Computing at the University of Utah are gratefully acknowledged. This research is also supported by National Artificial Intelligence Research Resource
(NAIRR) Pilot (award NAIRR240144) and the Delta advanced computing and data resource which is supported by the
National Science Foundation (award NSF-OAC
2005572). 

\bibliography{custom}

\begin{thebibliography}{18}
\providecommand{\natexlab}[1]{#1}

\bibitem[{Bansal et~al.(2024)Bansal, Hosseini, Agarwal, Tran, and Kazemi}]{bansal2024smaller}
Hritik Bansal, Arian Hosseini, Rishabh Agarwal, Vinh~Q Tran, and Mehran Kazemi. 2024.
\newblock Smaller, weaker, yet better: Training llm reasoners via compute-optimal sampling.
\newblock \emph{arXiv preprint arXiv:2408.16737}.

\bibitem[{Brown et~al.(2024)Brown, Juravsky, Ehrlich, Clark, Le, R{\'e}, and Mirhoseini}]{brown2024large}
Bradley Brown, Jordan Juravsky, Ryan Ehrlich, Ronald Clark, Quoc~V Le, Christopher R{\'e}, and Azalia Mirhoseini. 2024.
\newblock Large language monkeys: Scaling inference compute with repeated sampling.
\newblock \emph{arXiv preprint arXiv:2407.21787}.

\bibitem[{Dang et~al.(2024)Dang, Singh, D'souza, Ahmadian, Salamanca, Smith, Peppin, Hong, Govindassamy, Zhao et~al.}]{dang2024aya}
John Dang, Shivalika Singh, Daniel D'souza, Arash Ahmadian, Alejandro Salamanca, Madeline Smith, Aidan Peppin, Sungjin Hong, Manoj Govindassamy, Terrence Zhao, and 1 others. 2024.
\newblock Aya expanse: Combining research breakthroughs for a new multilingual frontier.
\newblock \emph{arXiv preprint arXiv:2412.04261}.

\bibitem[{Grattafiori et~al.(2024)Grattafiori, Dubey, Jauhri, Pandey, Kadian, Al-Dahle, Letman, Mathur, Schelten, Vaughan et~al.}]{grattafiori2024llama}
Aaron Grattafiori, Abhimanyu Dubey, Abhinav Jauhri, Abhinav Pandey, Abhishek Kadian, Ahmad Al-Dahle, Aiesha Letman, Akhil Mathur, Alan Schelten, Alex Vaughan, and 1 others. 2024.
\newblock The llama 3 herd of models.
\newblock \emph{arXiv preprint arXiv:2407.21783}.

\bibitem[{Guo et~al.(2025)Guo, Yang, Zhang, Song, Zhang, Xu, Zhu, Ma, Wang, Bi et~al.}]{guo2025deepseek}
Daya Guo, Dejian Yang, Haowei Zhang, Junxiao Song, Ruoyu Zhang, Runxin Xu, Qihao Zhu, Shirong Ma, Peiyi Wang, Xiao Bi, and 1 others. 2025.
\newblock Deepseek-r1: Incentivizing reasoning capability in llms via reinforcement learning.
\newblock \emph{arXiv preprint arXiv:2501.12948}.

\bibitem[{Gureja et~al.(2024)Gureja, Miranda, Islam, Maheshwary, Sharma, Winata, Lambert, Ruder, Hooker, and Fadaee}]{gureja2024m}
Srishti Gureja, Lester James~V Miranda, Shayekh~Bin Islam, Rishabh Maheshwary, Drishti Sharma, Gusti Winata, Nathan Lambert, Sebastian Ruder, Sara Hooker, and Marzieh Fadaee. 2024.
\newblock M-rewardbench: Evaluating reward models in multilingual settings.
\newblock \emph{arXiv preprint arXiv:2410.15522}.

\bibitem[{Lambert et~al.(2024)Lambert, Pyatkin, Morrison, Miranda, Lin, Chandu, Dziri, Kumar, Zick, Choi et~al.}]{lambert2024rewardbench}
Nathan Lambert, Valentina Pyatkin, Jacob Morrison, LJ~Miranda, Bill~Yuchen Lin, Khyathi Chandu, Nouha Dziri, Sachin Kumar, Tom Zick, Yejin Choi, and 1 others. 2024.
\newblock Rewardbench: Evaluating reward models for language modeling.
\newblock \emph{arXiv preprint arXiv:2403.13787}.

\bibitem[{Li et~al.(2024)Li, Chiang, Frick, Dunlap, Wu, Zhu, Gonzalez, and Stoica}]{li2024crowdsourced}
Tianle Li, Wei-Lin Chiang, Evan Frick, Lisa Dunlap, Tianhao Wu, Banghua Zhu, Joseph~E Gonzalez, and Ion Stoica. 2024.
\newblock From crowdsourced data to high-quality benchmarks: Arena-hard and benchbuilder pipeline.
\newblock \emph{arXiv preprint arXiv:2406.11939}.

\bibitem[{Lou et~al.(2024)Lou, Yan, Shen, Yan, Xie, and Zhang}]{lou2024uncertainty}
Xingzhou Lou, Dong Yan, Wei Shen, Yuzi Yan, Jian Xie, and Junge Zhang. 2024.
\newblock Uncertainty-aware reward model: Teaching reward models to know what is unknown.
\newblock \emph{arXiv preprint arXiv:2410.00847}.

\bibitem[{Madaan et~al.(2023)Madaan, Tandon, Gupta, Hallinan, Gao, Wiegreffe, Alon, Dziri, Prabhumoye, Yang et~al.}]{madaan2023self}
Aman Madaan, Niket Tandon, Prakhar Gupta, Skyler Hallinan, Luyu Gao, Sarah Wiegreffe, Uri Alon, Nouha Dziri, Shrimai Prabhumoye, Yiming Yang, and 1 others. 2023.
\newblock Self-refine: Iterative refinement with self-feedback.
\newblock \emph{Advances in Neural Information Processing Systems}, 36:46534--46594.

\bibitem[{Muennighoff et~al.(2025)Muennighoff, Yang, Shi, Li, Fei-Fei, Hajishirzi, Zettlemoyer, Liang, Cand{\`e}s, and Hashimoto}]{muennighoff2025s1}
Niklas Muennighoff, Zitong Yang, Weijia Shi, Xiang~Lisa Li, Li~Fei-Fei, Hannaneh Hajishirzi, Luke Zettlemoyer, Percy Liang, Emmanuel Cand{\`e}s, and Tatsunori Hashimoto. 2025.
\newblock s1: Simple test-time scaling.
\newblock \emph{arXiv preprint arXiv:2501.19393}.

\bibitem[{Singh et~al.(2024)Singh, Vargus, Dsouza, Karlsson, Mahendiran, Ko, Shandilya, Patel, Mataciunas, OMahony, Zhang, Hettiarachchi, Wilson, Machado, Moura, Krzemiński, Fadaei, Ergün, Okoh, Alaagib, Mudannayake, Alyafeai, Chien, Ruder, Guthikonda, Alghamdi, Gehrmann, Muennighoff, Bartolo, Kreutzer, Üstün, Fadaee, and Hooker}]{singh2024aya}
Shivalika Singh, Freddie Vargus, Daniel Dsouza, Börje~F. Karlsson, Abinaya Mahendiran, Wei-Yin Ko, Herumb Shandilya, Jay Patel, Deividas Mataciunas, Laura OMahony, Mike Zhang, Ramith Hettiarachchi, Joseph Wilson, Marina Machado, Luisa~Souza Moura, Dominik Krzemiński, Hakimeh Fadaei, Irem Ergün, Ifeoma Okoh, and 14 others. 2024.
\newblock \href {https://arxiv.org/abs/2402.06619} {Aya dataset: An open-access collection for multilingual instruction tuning}.
\newblock \emph{Preprint}, arXiv:2402.06619.

\bibitem[{Snell et~al.(2024)Snell, Lee, Xu, and Kumar}]{snell2024scaling}
Charlie Snell, Jaehoon Lee, Kelvin Xu, and Aviral Kumar. 2024.
\newblock Scaling llm test-time compute optimally can be more effective than scaling model parameters.
\newblock \emph{arXiv preprint arXiv:2408.03314}.

\bibitem[{Team(2024)}]{gemma_2024}
Gemma Team. 2024.
\newblock \href {https://doi.org/10.34740/KAGGLE/M/3301} {Gemma}.

\bibitem[{Wu et~al.(2024)Wu, Sun, Li, Welleck, and Yang}]{wu2024inference}
Yangzhen Wu, Zhiqing Sun, Shanda Li, Sean Welleck, and Yiming Yang. 2024.
\newblock Inference scaling laws: An empirical analysis of compute-optimal inference for problem-solving with language models.
\newblock \emph{arXiv preprint arXiv:2408.00724}.

\bibitem[{Yang et~al.(2024{\natexlab{a}})Yang, Yang, Zhang, Hui, Zheng, Yu, Li, Liu, Huang, Wei et~al.}]{yang2024qwen2}
An~Yang, Baosong Yang, Beichen Zhang, Binyuan Hui, Bo~Zheng, Bowen Yu, Chengyuan Li, Dayiheng Liu, Fei Huang, Haoran Wei, and 1 others. 2024{\natexlab{a}}.
\newblock Qwen2. 5 technical report.
\newblock \emph{arXiv preprint arXiv:2412.15115}.

\bibitem[{Yang et~al.(2024{\natexlab{b}})Yang, Ding, Lin, Zhang, and Zhang}]{yang2024regularizing}
Rui Yang, Ruomeng Ding, Yong Lin, Huan Zhang, and Tong Zhang. 2024{\natexlab{b}}.
\newblock Regularizing hidden states enables learning generalizable reward model for llms.
\newblock \emph{arXiv preprint arXiv:2406.10216}.

\bibitem[{Zheng et~al.(2023)Zheng, Chiang, Sheng, Zhuang, Wu, Zhuang, Lin, Li, Li, Xing et~al.}]{zheng2023judging}
Lianmin Zheng, Wei-Lin Chiang, Ying Sheng, Siyuan Zhuang, Zhanghao Wu, Yonghao Zhuang, Zi~Lin, Zhuohan Li, Dacheng Li, Eric Xing, and 1 others. 2023.
\newblock Judging llm-as-a-judge with mt-bench and chatbot arena.
\newblock \emph{Advances in Neural Information Processing Systems}, 36:46595--46623.

\end{thebibliography}

\appendix
\section{Other Experimental Details}
\label{sec:appendix:exp}
We provide some additional experimental details regarding the models and the hyperparameters used in our evaluation.

\subsection{Dataset and Languages.}
We use the Aya Evaluation Suite for conducting all our experiments~\cite{singh2024aya}.\footnote{\url{https://huggingface.co/datasets/CohereLabs/aya_evaluation_suite}} The dataset contains open-ended conversation-style, culturally grounded prompts for evaluating multilingual generation in LLMs. We demonstrate our results on nine languages, namely, Arabic (\texttt{ar}), English (\texttt{en}), Hindi (\texttt{hi}), Punjabi (\texttt{pa}), Portuguese (\texttt{pt}), Russian (\texttt{ru}), Telugu (\texttt{te}), Turkish (\texttt{tr}), and Chinese (\texttt{zh}). To get the Punjabi subset, we translate Hindi examples to Punjabi using Google Translate followed by manual post-editing. Second dataset is the m-ArenaHard dataset which is the multilingual variant of ArenaHard-Auto~\citep{li2024crowdsourced}. This dataset was released by ~\citet{dang2024aya} and contains the translated prompts from the ArenaHard-Auto benchmark. With this dataset, we perform evaluation in seven languages: Arabic (\texttt{ar}), Czech (\texttt{cs}), English (\texttt{en}), Hindi (\texttt{hi}), Japanese (\texttt{ja}), Portuguese (\texttt{po}), and Vietnamese (\texttt{vi}). 

\subsection{Verifiers}
\label{subsec:appendix:verifier}
We employ two types of verifiers to score generations produced through repeated sampling. For perplexity-based verification, we use the LLaMA-3.1-8B and Gemma-2B models. These models are used in their pre-trained forms, as they serve solely to compute perplexity scores.
For reward model-based scoring, we incorporate two classifier reward models that have been fine-tuned on top of these pre-trained architectures. The first is the \texttt{URM-LLaMA-3.1-8B} model~\footnote{\url{https://huggingface.co/LxzGordon/URM-LLaMa-3.1-8B}}, which achieves the highest overall score among all models with fewer than 8 billion parameters on the RewardBench benchmark~\cite{lambert2024rewardbench}.\footnote{\url{https://huggingface.co/spaces/allenai/reward-bench}} The second is the \texttt{GRM-Gemma-2B-rewardmodel-ft}~\footnote{\url{https://huggingface.co/Ray2333/GRM-Gemma-2B-rewardmodel-ft}}, a reward model fine-tuned on the Gemma-2B backbone, and the leading performer in the sub-3B parameter category on RewardBench. Using reward models that share the same backbone as the perplexity models helps reduce confounding factors and allows for clearer interpretation of results in our experimental comparisons.

It is important to note that both reward models are trained exclusively on English-language data. We deliberately choose not to explore multilingual reward models for two main reasons. First, the best-performing open-weight multilingual reward models, such as Gemma-27B, are significantly larger in scale, which limits their practicality in repeated sampling scenarios. Second, the objective of this work is to demonstrate that simple, readily available verifiers can already yield meaningful improvements in multilingual generation. That said, developing a smaller yet effective multilingual reward model remains an interesting direction for future research. As an aside, prior work has shown that the English-only \texttt{URM-LLaMA-3.1-8B} model can be effective across multiple languages~\cite{gureja2024m}.

\subsection{Models and Hyperparameters.}

We evaluate the most common multilingual, open-weight LLMs at both their large and small parameter scales. Specifically, we use Qwen-2.5~\cite{yang2024qwen2} at 14B, 72B parameter scales, Llama-3 at 8B and 70B scales~\cite{grattafiori2024llama}~\footnote{We use Llama-3.1 for the 8B model, and Llama-3.3 for the 70B model. Llama-3.3 does not have a corresponding 8B model and is known to be much superior to the Llama-3.1 70B model.}, and aya-expanse at 8B and 32B~\cite{dang2024aya}. We abbreviate these names in figures to make them more readable.

\paragraph{Hyperparameters.} During our preliminary experiments, we found that temperature and other sampling parameters like top-p did not degrade the generation quality. We therefore fixed the temperature to 0.8, and top-p to 0.95. 

\paragraph{GPT-4o vs Gemini for Judge.} We used Gemini's 2.0 flash model for calculating win/loss rates. For a subset of our experiments on Aya Evaluation Suite with Llama-based verifiers, we found that on average the delta scores from GPT-4o and Gemini were within 3.0\% of each other. Gemini model is 25x cheaper than the GPT-4o model and therefore considering the scale of our evaluation, it made more sense to use the Gemini model.

\begin{figure}[t]
    \centering
    \includegraphics[width=\columnwidth]{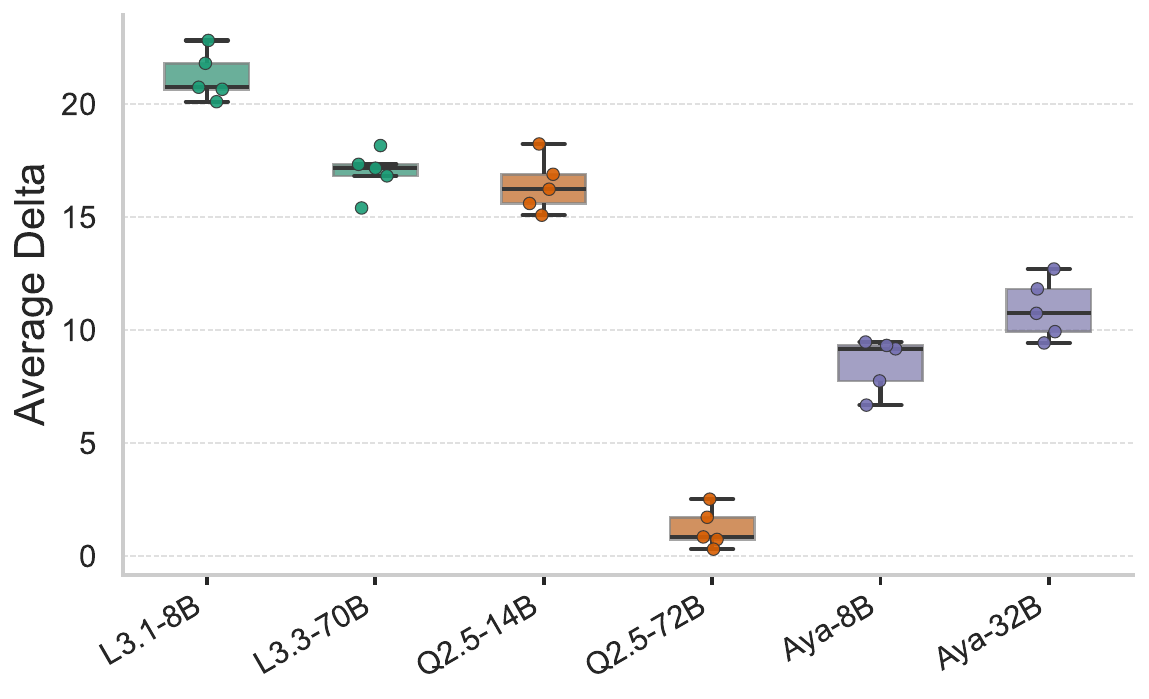}
    \caption{Impact of baseline used for win rate calculation. We use \texttt{gemini-2.0-flash} as the judge model and evaluate stability with a reward model as a verifier: \texttt{URM-LLaMa-3.1-8B}}
    \label{fig:stability}
\end{figure}
\paragraph{What baseline response to use?}
We need to compare the response chosen by the verifier with a baseline sampled response (i.e. when there is no repeated sampling). Due to randomness in sampling, this comparison itself becomes non-deterministic. Of course, ideally, we want to perform the direct comparison with multiple different baseline responses. But this is not practical due to budget constraints. 

We find that this does not have a significant impact. See~\cref{fig:stability} for a box plot that shows the variability of the delta scores when comparing it with different baseline responses. Although there is some variance, but as can be clearly seen, it is within a small range (around 2 \% points). 

\section{Detailed Results and Plots}
\begin{figure*}[ht]
    \centering
    \includegraphics[width=\textwidth]{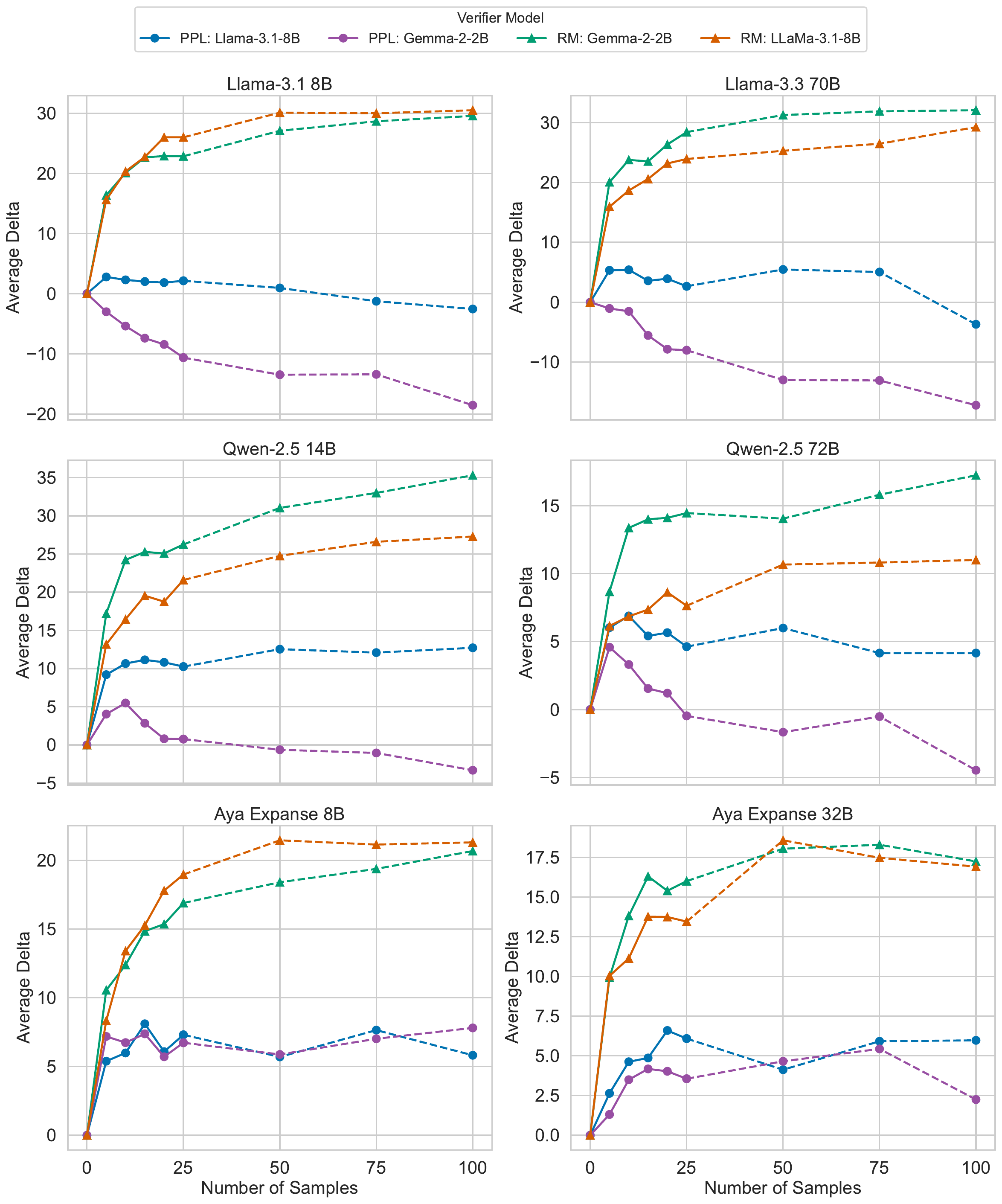}
    \caption{Inference time scaling with repeated sampling for multilingual generation. Results for the m-ArenaHard.}
    \label{fig:scaling_main_arena_all}
\end{figure*}
\begin{figure*}[ht]
    \centering
    \includegraphics[width=\textwidth]{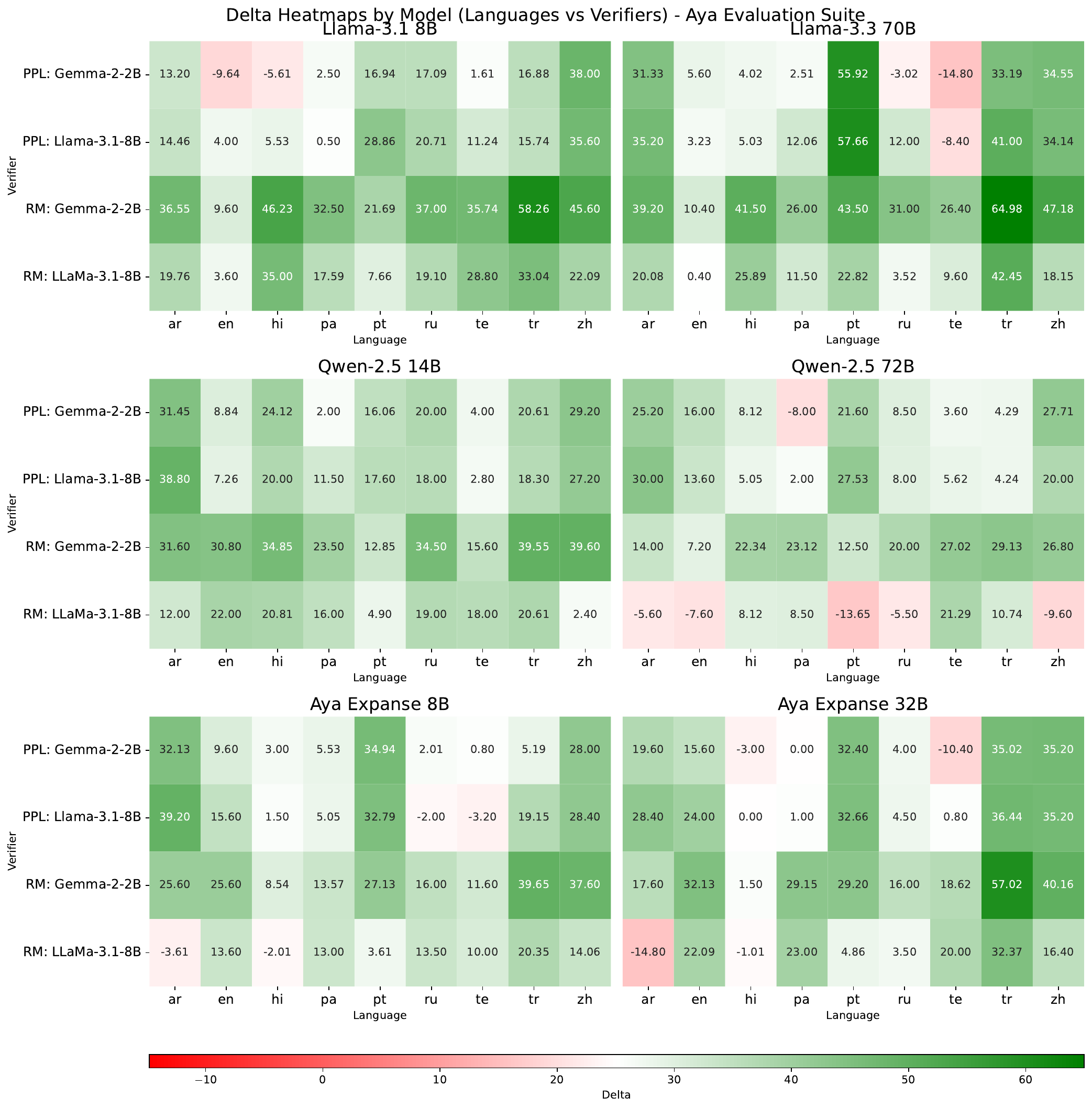}
    \caption{For Aya Evaluation Suite. Heatmaps showing the language-specific results for each model and verifier. The results are shown for $n = 100$. }
    \label{fig:heatmap_aya}
\end{figure*}

\begin{figure*}[ht]
    \centering
    \includegraphics[width=\textwidth]{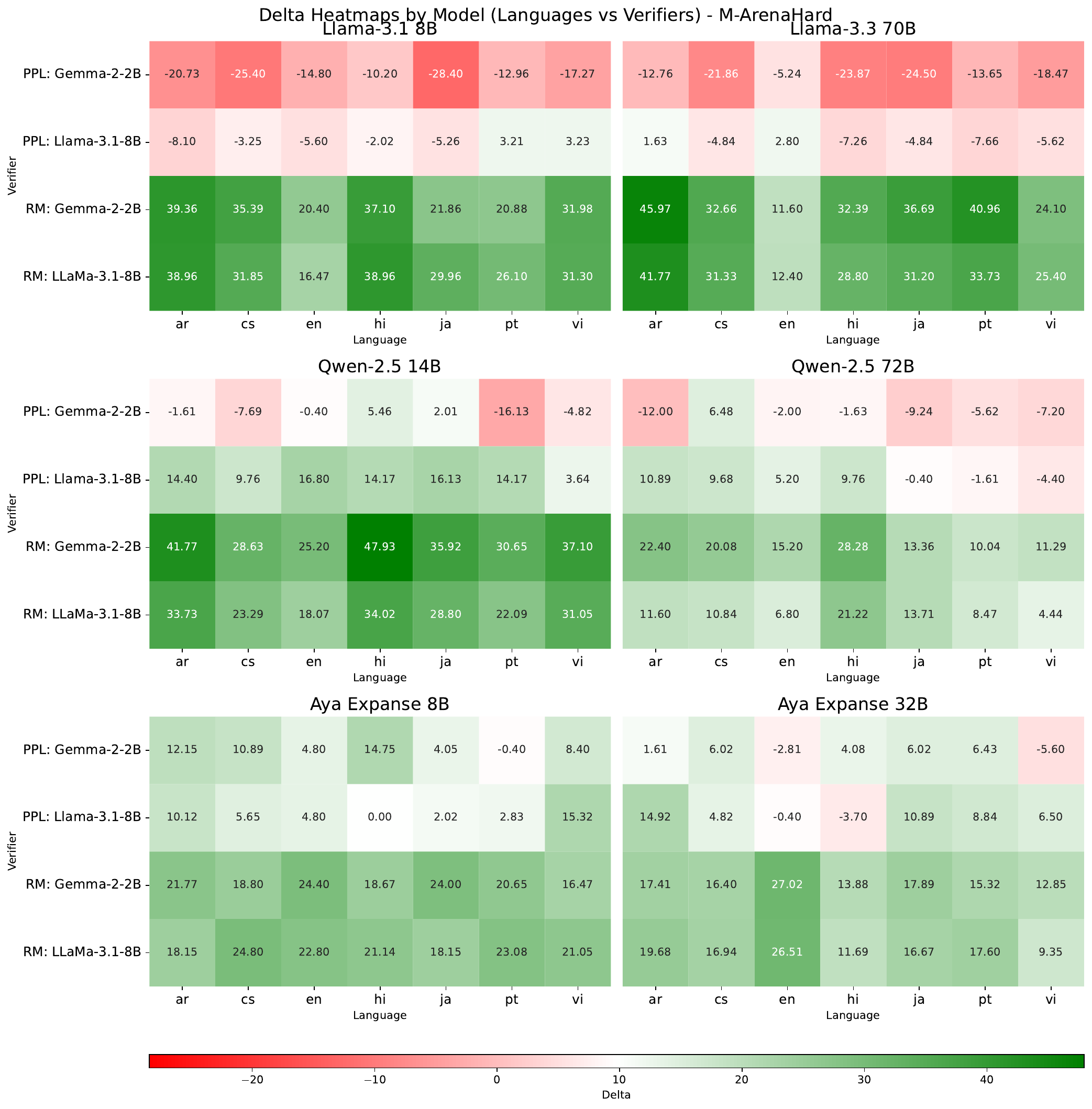}
    \caption{For m-ArenaHard. Heatmaps showing the language-specific results for each model and verifier. The results are shown for $n = 100$. }
    \label{fig:heatmap_arena}
\end{figure*}

\end{document}